%% file: conference_101719.tex
\documentclass[conference]{IEEEtran}
\IEEEoverridecommandlockouts
\usepackage[hidelinks, bookmarks=false]{hyperref}
\usepackage{cite}
\usepackage{amsmath,amssymb,amsfonts}
\usepackage{algorithmic}
\usepackage{graphicx}
\usepackage{textcomp}
\usepackage{xcolor}
\usepackage{comment}
\usepackage{mathtools}
\usepackage{amsmath}
\usepackage{amssymb}
\usepackage{mathtools}
\usepackage{amsthm}
\usepackage{fixmath}
\usepackage{amsmath}
\usepackage{amsfonts}
\usepackage{amssymb}
\usepackage{multicol}
\usepackage{multirow}
\usepackage{tabularx,ragged2e}
\usepackage{booktabs}
\newcolumntype{L}{>{\centering\arraybackslash}X}

\def\BibTeX{{\rm B\kern-.05em{\sc i\kern-.025em b}\kern-.08em
    T\kern-.1667em\lower.7ex\hbox{E}\kern-.125emX}}
\begin{document}

\title{Efficient Noise Mitigation for Enhancing Inference Accuracy in DNNs on Mixed-Signal Accelerators
}
\author{\IEEEauthorblockN{Seyedarmin Azizi*, Mohammad Erfan Sadeghi*\thanks{*Seyedarmin Azizi and Mohammad Erfan Sadeghi contributed equally to this work.}, Mehdi Kamal, and Massoud Pedram}
\IEEEauthorblockA{Department of Electrical \& Computer Engineering, University of Southern California, Los Angeles, CA, USA\\
    \url{{seyedarm,sadeghim,mehdi.kamal,pedram}@usc.edu}}
}
\maketitle

\begin{abstract}
In this paper, we propose a framework to enhance the robustness of the neural models by mitigating the effects of process-induced and aging-related variations of analog computing components on the accuracy of the analog neural networks. We model these variations as the noise affecting the precision of the activations and introduce a denoising block inserted between selected layers of a pre-trained model. We demonstrate that training the denoising block significantly increases the model's robustness against various noise levels. To minimize the overhead associated with adding these blocks, we present an exploration algorithm to identify optimal insertion points for the denoising blocks. Additionally, we propose a specialized architecture to efficiently execute the denoising blocks, which can be integrated into mixed-signal accelerators. We evaluate the effectiveness of our approach using Deep Neural Network (DNN) models trained on the ImageNet and CIFAR-10 datasets. The results show that on average, by accepting 2.03\% parameter count overhead, the accuracy drop due to the variations reduces from 31.7\% to 1.15\%.


\end{abstract}


\section{Introduction}
\input{introduction}

\section{Background}
\subsection{Denoising in Deep Learning}

\input{denoisinglayer}
\subsection{Analog MVM Reliability}

\input{MVMreliability}
\subsection{Related work}
\input{relatedwork}

\section{Methodology}
In subsection A, we provide an overview of the denoising mechanism and the mathematical foundation of our framework. In subsection B, we describe the proposed denoising block. Finally, in subsection C, we explain how to determine the optimal placement of the denoising block in a neural network to balance its noise reduction effectiveness with computational overhead. Throughout this section, we denote the given pretrained neural network model (without any noise) as \(\mathcal{M}\), and the noisy version which is also equipped with the denoising module, as \(\mathcal{M}^*\). We use \(\mathcal{L}\) as the neural network loss and the \(\mathcal{D}\) as our dataset, which contains the input data samples and the ground truth. Finally, in subsection D, we present a detailed analysis of the denoising block's hardware architecture along with its data flow.

\subsection{Denoising Overview}
\input{denoising_overview}

\subsection{Denoising block}
\input{denoiser_block}

\subsection{Integration of Denoising Block}
\input{adding_denoiser}

\subsection{Denoiser Hardware Architecture}
\input{denoiser_hw}

\begin{figure}[tb] 
    \centering
     \includegraphics[scale=0.5]{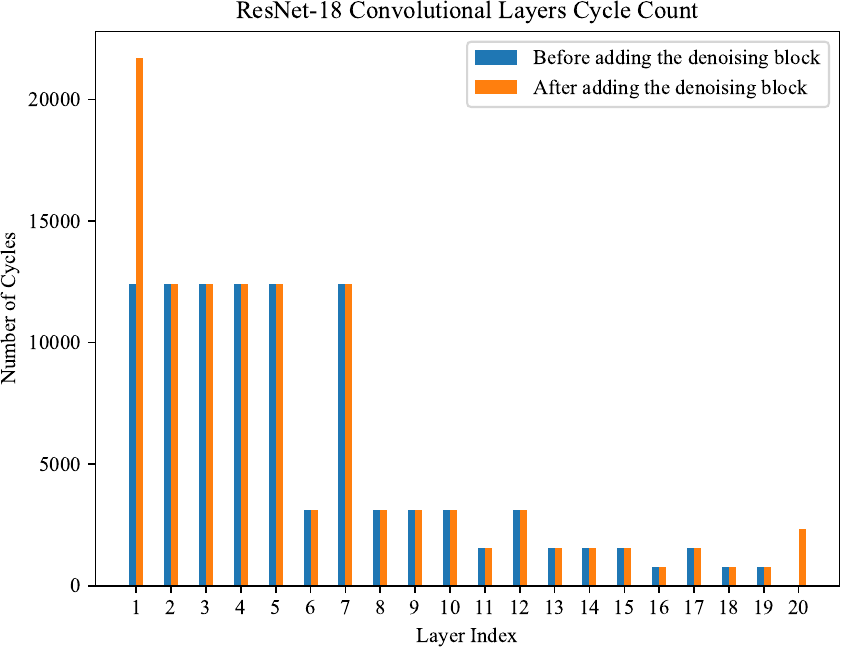}

    \caption{Cycle count for ResNet-18 layers}
    \label{fig:latency}

\end{figure}

\section{Results and Discussion}
In this section we first provide the experimental setup of our framework, then we detail our main results on the denoising performance and the hardware metrics, including power. 
\input{results}







\section{Conclusion}
\input{conclusion}

\bibliographystyle{IEEEtran}
\bibliography{IEEEabrv,conference_101719}
	
\end{document}

%% file: introduction.tex
Deep Neural Networks (DNNs) \cite{DBLP:conf/cvpr/HeZRS16, DBLP:journals/cacm/KrizhevskySH17, DBLP:journals/corr/VaswaniSPUJGKP17} have become a cornerstone in the field of artificial intelligence and machine learning due to their remarkable ability to model complex data patterns and perform sophisticated tasks. DNNs excel in learning hierarchical representations of data through multiple layers of abstraction, enabling them to automatically extract features and generalize well across a wide variety of tasks such as image recognition, natural language processing, and decision-making in complex environments such as robotics.


While Deep Neural Networks (DNNs) deliver exceptional performance, they incur significant computational costs. Both training and inference require substantial processing power due to the large number of parameters and operations, including both linear and non-linear computations \cite{livni2014computational, han2015deep, sadeghi2024peano, lecun2002efficient}. Specialized hardware accelerators are essential for efficiently executing these models. Traditional CPUs lack the necessary parallelism for DNN workloads, while GPUs excel in parallel processing but consume high power, limiting their use in energy-constrained environments like edge devices. FPGAs \cite{mittal2020survey, erfan2024chosen, guan2017fpga, zhang2015optimizing}, although customizable hardware accelerators, are fully digital and tend to have relatively high latency compared to GPUs. Additionally, their power consumption, while lower than GPUs in some cases, is still significant, making them less ideal for certain applications requiring extreme power efficiency.

While some accelerators are fully digital, mixed-signal accelerators present a promising alternative, particularly for matrix-vector multiplication (MVM) \cite{shafiee2016isaac}, a critical operation in DNNs. Analog computing \cite{9252918} is well-suited for MVM, offering significant gains in power efficiency and computational speed. Techniques explored include current-based methods with MOSFETs, charge-based approaches using SRAMs \cite{zhang2022177}, and resistance-based technologies like memristors \cite{10.1145/3297858.3304049, DBLP:journals/corr/PreziosoMHALS14} and phase-change memory (PCM) \cite{10035378}. However, analog computations face reliability challenges due to process variations during manufacturing and temporal variations over time, which reduce computational accuracy. Addressing these issues to ensure precision and stability in analog accelerators is a key focus of ongoing research.

To enhance the reliability of analog-based computations, several techniques have been developed. One common approach is training-based methods \cite{9533684, 9472868}, where models are trained to tolerate lower-precision operations, thereby increasing their robustness to variations in analog hardware. 
Additionally, error correction codes (ECC) \cite{lou2020embedding, 9174137} are used to detect and correct errors introduced during analog computations or data transmission, further improving reliability. Innovations at the device level, such as using more robust materials and advanced fabrication processes, help mitigate the effects of process and temporal variations. Furthermore, stochastic computing \cite{8746640, 7093194} offers a probabilistic approach that inherently tolerates errors by relying on statistical properties of computations, thus enhancing reliability.

In this work, we introduce a probabilistic denoising module that enhances the robustness of DNNs with minimal hardware overhead. Unlike traditional methods that require retraining the entire model, our approach, inspired by parameter-efficient fine-tuning literature \cite{hu2021lora, azizi2024lamda, hu2023llm} freezes the main model and trains only the denoising block, utilizing parameter-efficient fine-tuning to reduce time, memory, and computational costs. We also propose an algorithm to determine the optimal placement of the denoising blocks within the network. This flexible method is applicable to any analog-based MVM architecture, addressing reliability challenges from noise and variations in analog hardware. Finally, we present the architecture of the denoising block to demonstrate the feasibility of our approach.

%% file: denoisinglayer.tex
The task of denoising, or recovering clean signals from corrupted observations, has long been a central problem in signal and image processing. Early approaches, such as Wiener filtering \cite{wiener1949extrapolation} and total variation minimization \cite{Rudin1992NonlinearTV}, operated based on predefined statistical assumptions about the signal and noise. However, these methods often struggled with real-world complexities, where noise characteristics are not easily modeled by simple assumptions. Deep learning methods, particularly convolutional neural networks (CNNs), have drastically improved denoising performance by learning noise patterns directly from data. A notable example is DnCNN \cite{DBLP:journals/tip/ZhangZCM017}, which predicts the residual noise in an image and subtracts it to produce a clean result. This residual learning strategy is effective but computationally expensive due to the use of large convolutional layers for denoising. Autoencoders \cite{vincent2008extracting} and generative models like GANs \cite{goodfellow2020generative} have also been used, with GAN-based denoising \cite{yang2017deep} leveraging adversarial training to generate realistic clean images. However, these methods typically lack training and inference efficiency.

Probabilistic models, such as Variational Autoencoders (VAEs) \cite{kingma2013auto} and score-based generative models \cite{DBLP:conf/iclr/0011SKKEP21}, model noise by capturing the uncertainty inherent in noisy data. These models estimate not just a clean signal but a distribution over possible clean signals, offering robustness in complex noise environments. 
Diffusion models, such as Denoising Diffusion Probabilistic Models (DDPM) \cite{ho2020denoising}, offer a probabilistic framework for denoising by iteratively adding noise in a forward process and learning to reverse this process to recover the original data. The key strength of diffusion models is their stepwise refinement of noisy inputs, allowing for precise noise removal over several iterations. While highly effective, diffusion models are computationally intensive, as they require multiple passes to gradually denoise the signal. In our bottleneck-based denoising block, we take inspiration from diffusion models' probabilistic reasoning by predicting both the mean and variance of the noise in a \textbf{single pass}. 

Bottleneck architectures, widely used in networks like ResNets \cite{he2016deep}, reduce the computational cost of deep models by compressing intermediate representations into lower-dimensional spaces. Depthwise separable convolutions \cite{DBLP:conf/cvpr/Chollet17} extend this by separating the spatial and channel-wise operations, significantly reducing the number of parameters and computations required.


%% file: MVMreliability.tex
Analog-based matrix-vector multiplication (MVM) accelerators have gained attention for their potential to significantly improve power efficiency and computational speed in deep neural networks (DNNs).  These accelerators utilize analog computing to perform MVM operations, a fundamental component in both DNN inference and training. By exploiting the continuous properties of analog circuits, they achieve reductions in both the energy required for computation and the on-chip area. Despite these promising advantages, analog MVM accelerators face critical challenges related to reliability, particularly due to noise, process variations, and temporal variations in analog components.

A key issue affecting the reliability of analog MVM accelerators is process variations that occur during manufacturing, leading to inconsistencies in device performance for components such as memristors, phase-change memory (PCM), and MOSFETs. These devices, which rely on properties like current flow, resistance, and charge storage, are highly susceptible to fabrication deviations. Minor manufacturing inconsistencies can introduce errors in analog computations, diminishing the precision of MVM operations and degrading overall system reliability. Additionally, temporal variations—caused by environmental factors like temperature fluctuations and voltage drift—can further degrade analog component performance over time, resulting in shifting operational characteristics and reduced computational accuracy. Unlike digital systems, which are resilient to minor changes, analog systems are highly vulnerable to errors induced by these variations, exacerbating reliability issues.

The increased noise and reduced precision in MVM operations significantly affect the inference accuracy of DNNs, especially as their scale and complexity grow, leading to degraded overall system performance. In high-accuracy applications, the unreliability of analog MVM accelerators becomes particularly problematic. Given the critical role of matrix-vector multiplications in DNN performance and the growing interest in mixed-signal systems for edge computing and energy efficiency, addressing the reliability challenges in analog MVM accelerators is essential. Ensuring reliable operation despite process and temporal variations is crucial for maintaining high inference accuracy and enabling the practical deployment of energy-efficient, high-precision DNNs in real-world applications.

%% file: relatedwork.tex
\subsubsection{Training-based methods}

Previous work, such as \cite{9533684} and \cite{9472868}, has proposed noise-aware training methods for DNNs. While these methods enhance accuracy by enabling the model to adapt to noisy environments, they require retraining the entire model from scratch, which is particularly time-consuming and computationally expensive, especially for larger DNNs. Furthermore, these approaches often employ low-precision computations to increase the model’s robustness to noise, which can negatively affect overall performance.

\subsubsection{Stochastic computing based methods}

While stochastic computing (SC) offers advantages in energy efficiency and simplicity, it also presents several challenges. One of the primary issues is the inherent loss of precision due to the use of probabilistic bitstreams to represent data, which often results in accuracy trade-offs, especially in more complex neural networks. To achieve higher precision, SC requires longer bitstreams, which can slow down computations and reduce the overall efficiency benefits. Additionally, SC typically necessitates retraining models from scratch, as its computation methods differ fundamentally from traditional binary systems, further increasing computational cost and time \cite{8746640, 7093194}.

\subsubsection{ECC-based methods}

Error Correction Code (ECC) methods, such as those in \cite{lou2020embedding} and \cite{9174137}, are commonly used in deep neural networks (DNNs) to enhance robustness by detecting and correcting errors during computation or data transmission, especially in noisy analog and mixed-signal systems. However, ECC has a limited error correction capacity, typically handling only small, isolated errors like single-bit or double-bit errors. In high-noise environments, its ability to detect and correct errors becomes inadequate, leading to accumulated uncorrected errors and reduced system reliability. Additionally, like training-based and stochastic methods, ECC often necessitates retraining the model to adapt to the changes from ECC encoding.



%% file: denoising_overview.tex
The denoising process is fundamentally concerned with estimating and mitigating the noise component embedded within a noisy input tensor \(\mathbold{X}\). Specifically, given \(\mathbold{X}\), which contains an unknown noise component \(\mathbold{Z}\) (where the clean signal is represented as \(\mathbold{X} - \mathbold{Z}\)), the goal is to predict a denoised output \(\mathbold{\hat{X}} = \mathbold{X} - \mathbold{\hat{Z}}\). Here, \(\mathbold{\hat{Z}}\) serves as an approximation of the true noise \(\mathbold{Z}\). Our denoising framework, inspired by principles from denoising diffusion probabilistic models (DDPMs) \cite{ho2020denoising}, assumes that the noise tensor \(\mathbold{Z}\) follows a Gaussian distribution, \(\mathbold{Z}_i \sim \mathcal{N}(\mu_i, \sigma_i^2)\), where each element of the noise is independently and identically distributed according to a Gaussian distribution characterized by a mean \(\mu_i\) and variance \(\sigma_i^2\).
The task of denoising can be formalized as a minimization of the expected reconstruction error between the unknown clean signal \(\mathbold{X} - \mathbold{Z}\) and the predicted denoised signal \(\mathbold{\hat{X}} = \mathbold{X} - \mathbold{\hat{Z}}\). Mathematically, this objective can be expressed as:

\begin{equation}
\begin{aligned}
\min_{\hat{\mu}, \hat{\sigma}^2} \mathbb{E}_{\mathbold{Z}, \mathbold{\hat{Z}}} \|\mathbold{\hat{X}} - (\mathbold{X} - \mathbold{Z})\|_2^2
= \min_{\hat{\mu}, \hat{\sigma}^2} \mathbb{E}_{\mathbold{Z}, \mathbold{\hat{Z}}} \|\mathbold{Z} - \mathbold{\hat{Z}}\|_2^2 \\
= \min_{\hat{\mu}, \hat{\sigma}^2} \mathbb{E}_{\epsilon} \|\mu + \sigma\epsilon - (\hat{\mu} + \hat{\sigma} \epsilon)\|_2^2,
\end{aligned}  
\end{equation}

\noindent where \(\epsilon \sim \mathcal{N}(\mathbold{0}, \mathbold{I})\) represents a standard normal variable, and \(\hat{\mu}\) and \(\hat{\sigma}^2\) are the predicted mean and variance of the noise, respectively. This formulation aims to ensure that the predicted noise closely approximates the true noise component by directly minimizing the discrepancy between their statistical characteristics.

The essence of the denoising task lies in accurately predicting the mean \(\hat{\mu}\) and variance \(\hat{\sigma}^2\) of the noise, as these parameters allow the reconstruction of an estimate of the noise \(\mathbold{\hat{Z}}\). This approach leverages the Gaussian assumption to model the noise structure, making it feasible to utilize simple yet effective probabilistic estimations. The denoising block leverages these predictions to sample noise as:

\begin{equation}
\label{eq:Z}
\mathbold{\hat{Z}} = \epsilon \odot \sqrt{\mathrm{Noise\_Var}} + \mathrm{Noise\_Mean},
\end{equation}

\noindent where the predicted variance, \(\mathrm{Noise\_Var} = \hat{\sigma}^2\), specifies the variability, and \(\mathrm{Noise\_Mean} = \hat{\mu}\) centers the noise around the predicted mean value. his probabilistic formulation aligns with the operation of denoising diffusion probabilistic models (DDPMs); however, instead of iteratively refining the estimate through multiple denoising steps, we achieve denoising in a\textbf{ single pass}, significantly reducing the computational burden. DDPMs employ a U-Net architecture to predict the mean $(\hat{\mu})$ at each denoising step, systematically removing it from the noisy input to create a less noisy version of the signal. This iterative approach is primarily designed for generative AI tasks, where the quality of denoising is crucial for image synthesis. In contrast, our objective goes beyond the denoising accuracy and focuses on enhancing the overall performance of the neural network for its specific task, particularly under the constraints of mixed-signal hardware that is susceptible to noise.
Instead of directly minimizing the denoising error alone, our framework thus integrates the denoising block within the neural network to minimize the task-specific loss \(\mathcal{L}\). The denoising parameters \(\hat{\mu}\) and \(\hat{\sigma}^2\) are optimized to reduce the expectation of the loss at the output of the network \(\mathcal{M}^*\):
\begin{equation}
\min_{\hat{\mu}, \hat{\sigma}} \mathbb{E}_{\epsilon} \mathcal{L}(\mathcal{M}^*_{\hat{\mu}, \hat{\sigma}}(x), y), \quad \forall x, y \in \mathcal{D},
\end{equation}
\noindent where the expectation is taken over the randomness introduced by the variable \(\epsilon\). This formulation underscores the primary innovation of our approach: treating the denoising block as an integrated component of the model rather than an isolated preprocessing step. 

The overall process involves the following steps:
(1) Noise Simulation: The pre-trained model \(\mathcal{M}\) is modified by introducing noise into its operations, with parameters \(\mu\) and \(\sigma\) judiciously chosen to simulate the non-idealities associated with the target mixed-signal hardware.
(2) Denoiser Integration: The denoising block is inserted into the model, resulting in the modified architecture \(\mathcal{M}^*\).
(3) Training the Denoiser: Parameters of the denoising block are trained, while \textbf{parameters of the original model \(\mathcal{M}\) remain fixed}. This ensures that the computational overhead is minimized, as only a small fraction of the model’s total parameters are updated. The assumption that the denoiser's operations are noise-free further simplifies the optimization process, allowing rapid convergence even with a limited number of epochs. In scenarios where hardware-induced noise disrupts model accuracy, fine-tuning \(\hat{\mu}\) and \(\hat{\sigma}^2\) provides a critical layer of correction that stabilizes the performance of the neural network. By embedding the denoiser within the model, our method effectively compensates for both process and temporal variations inherent in mixed-signal systems.
\begin{figure}[tb]
    \centering
    \includegraphics[width=\columnwidth]{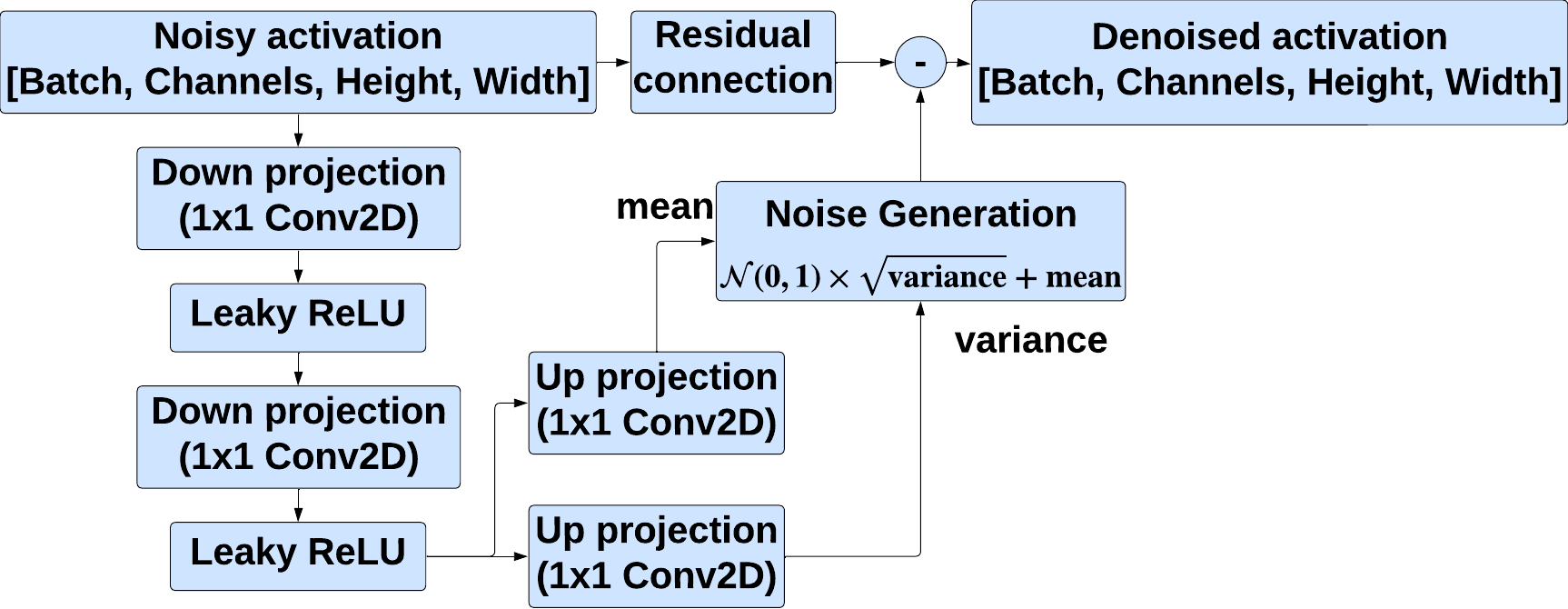}

    \caption{The proposed denoising block.}
    \label{fig:denoiser}

\end{figure}

\begin{figure*}[tb]
    \centering
     \includegraphics[scale=0.5]{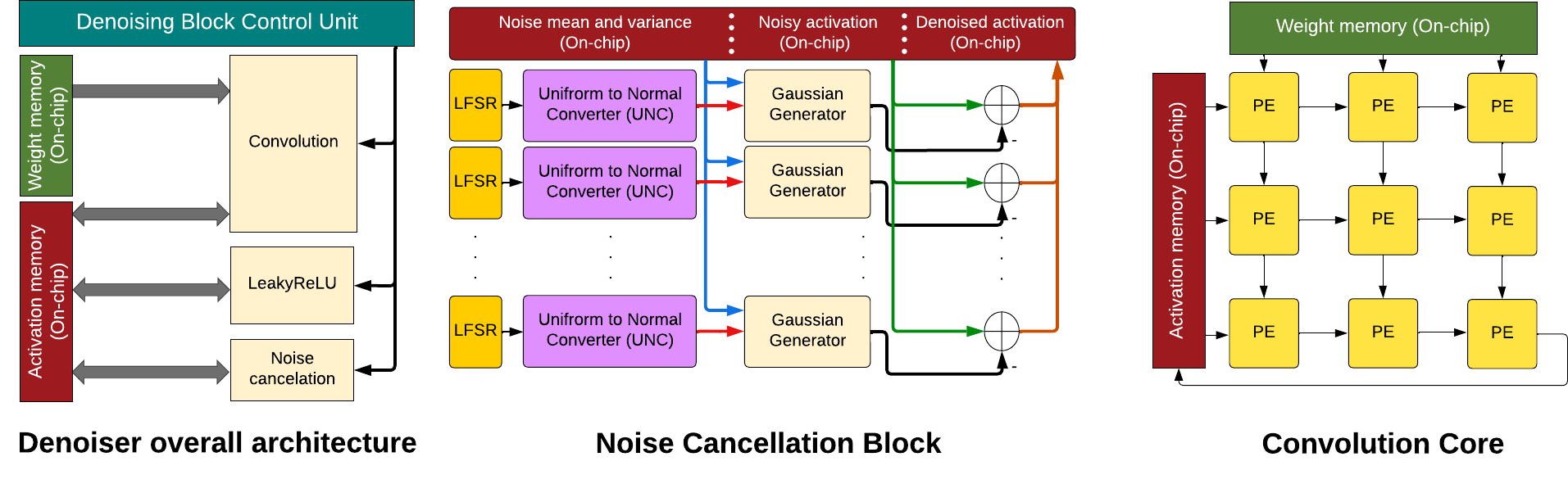}

    \caption{The denoising block hardware architecture}
    \label{fig:denoiser_hw}

\end{figure*}

%% file: denoiser_block.tex
\label{ref:denoiser_block}
The overall architecture of the proposed denoising block is shown in Fig. \ref{fig:denoiser}. The architecture borrows ideas from the bottleneck layer (block) found in neural network architectures like autoencoders and specific kinds of D models. The block processes the noisy input tensor \(\mathbold{X}\) by first applying a point-wise convolution (\(1 \times 1\) convolution) to reduce the channel dimension of the tensor. This operation is computationally efficient and parameter-light due to the use of a \(1 \times 1\) filter, which performs channel-wise transformation without altering the spatial dimensions.

Following this, a depth-wise convolution with a kernel size of \(3 \times 3\) is applied to capture spatial features in the reduced-dimensional representation. To ensure that the predicted noise mean and variance tensors match the dimensions of the input, two parallel point-wise convolutions are subsequently employed. These convolutions are responsible for generating the noise mean  and noise variance, while restoring the channel dimensions back to that of the original input. Given that the noise in each element of the input is assumed to follow a normal distribution, the noise estimation can be derived by sampling from a standard normal, scaling it by the predicted variance, and shifting it by the predicted mean according to the equation \ref{eq:Z} to obtain \(\hat{\mathbold{Z}}\). The denoised output is then obtained by subtracting the estimated noise from the input:
\begin{equation}
\hat{\mathbold{X}} = \mathbold{X} - \hat{\mathbold{Z}}
\end{equation}
\noindent A key advantage of the proposed denoising block is its lightweight design, which aligns with the principles of bottleneck architectures. This structure allows the module to be trained with minimal computational overhead, especially since the parameters of the original pretrained model remain fixed during training. This efficient integration ensures that the addition of the denoiser does not significantly impact the overall model’s computational cost.

%% file: adding_denoiser.tex
\label{sec:model_arch}
Integrating the denoising block after every convolution operation would result in significant computational and latency overheads, making this approach impractical. Therefore, to achieve a balance between denoising effectiveness and computational efficiency, we strategically insert the denoising block after a select few layers of the model. This selective placement aims to optimize performance while minimizing the additional overhead introduced by the denoiser.
    
Given a fixed budget \(\eta\) representing the percentage of the denoiser parameters to be utilized, our objective is to design an algorithm that optimally selects the layers where the denoising block should be inserted. Intuitively, the denoising block should be placed after layers whose outputs significantly impact the overall loss of the neural network. To formalize this intuition, we consider the first-order Taylor expansion of the neural network's loss function. Assuming noise is added only to the output of layer \(l\) while all other layers remain unchanged, the loss \(\mathcal{L}\) can be approximated as:

\begin{equation}
    \label{eq1}
    \mathcal{L}(\mathbold{y}_l + z) \approx \mathcal{L}(\mathbold{y}_l) + z^\mathrm{T} \mathbold{\nabla}_{\mathbold{y}_l} \mathcal{L}
\end{equation}

\noindent where \(\mathbold{y}_l\) denotes the output of layer \(l\), \(z\) represents the noise, and \(\mathbold{\nabla}_{\mathbold{y}_l} \mathcal{L}\) is the gradient of the loss with respect to the layer’s output. This approximation indicates that, for a given noise magnitude, the impact of noise on the model’s loss is proportional to the gradient norm of the layer’s output feature map. Consequently, layers with the highest gradient norm \(\|\mathbold{\nabla}_{\mathbold{y}_l} \mathcal{L}\|\) will have the most significant effect on the model's predictions. Therefore, the denoising block should be strategically placed after such layers to maximize its effectiveness while adhering to the parameter budget \(\eta\).

Based on this metric, we propose a heuristic approach to strategically insert the denoising blocks into the DNN model. Specifically, we compute the gradient of the loss with respect to the output feature maps of each layer in the pre-trained model and rank these gradients in descending order of their magnitudes. Denoising blocks are then incrementally added to the layers according to this sorted list, starting with those layers whose gradients have the highest magnitude, until the total parameter count reaches the specified budget \(\eta\).

%% file: denoiser_hw.tex
The hardware architecture of the implemented denoising block is depicted in Fig. \ref{fig:denoiser_hw}. We implemented the denoising block within a digital module to ensure that it operates free from noise and reliability issues, as it is essential for the denoiser itself to remain noise-free.

\subsubsection{Overall architecture}

The denoiser hardware architecture features a Denoiser Control Unit (DCU), which orchestrates the activation of specific computational blocks, including convolution, leaky ReLU, and noise cancellation modules. The DCU ensures that only the relevant block is activated during each phase of computation. Once a block is enabled, it retrieves the required data—such as weights and activations—from memory, executes the necessary operations, and stores the computed results back into memory upon completion.
To enhance the speed of the denoising block, we utilized on-chip memory for storing weights and activations, thereby reducing the overhead associated with external memory accesses. Additionally, we separated the weight memory from the activation memory, allowing the convolution block to simultaneously read both weight and activation values. This design optimizes data flow, reduces latency, and ensures the denoiser operates efficiently with minimal delays.

\subsubsection{Noise cancellation architecture}
The noise cancellation block is responsible for generating noise based on the predicted mean and variance, and subtracting it from the input—a process referred to as denoising or noise cancellation. To implement this, we employed the Box-Muller transform, as shown below:
\begin{equation} \label{box-muller}
\begin{gathered}
Z_1 = \sqrt{-2ln(U_1)}\times \cos{(2\pi U_2)} \\
Z_2 = \sqrt{-2ln(U_1)}\times \sin{(2\pi U_2)}
\end{gathered}
\end{equation}
The random variables \( U_1 \) and \( U_2 \) are independent and uniformly distributed, drawn from a \( \text{uniform}(0, 1) \) distribution, and \( Z_1 \) and \( Z_2 \) are normally distributed random variables. To generate these uniform random variables, we utilize multiple Linear Feedback Shift Registers (LFSRs), each initialized with distinct seed values. These LFSRs produce random outputs, which are normalized by dividing by the maximum representable value in the system's numerical format.

The normalized outputs of the LFSRs are then passed to Uniform-to-Normal Converters (UNCs), where the normally distributed outputs are generated from the uniform inputs using the Box-Muller transform (equation \ref{box-muller}). Since the sum \( Z_1^2 + Z_2^2 \) is constant, we utilize only \( Z_1 \) to ensure that the generated values remain fully independent.
Finally, the Gaussian noise generator uses the generated normal values, along with the predicted noise mean and variance, to produce the predicted noise values based on equation \ref{gaussian-from-normal}.

\begin{equation} \label{gaussian-from-normal}
\begin{gathered}
Z_1 \sim \mathcal{N}(0, 1) \quad and \quad Y = Z_1 \times \sigma + \mu \\
Y \sim Guassian(\mu, \sigma^2)
\end{gathered}
\end{equation}
Finally, the generated noise values are subtracted from the noisy activations to obtain the denoised activations.

 \subsubsection{Convolution core architecture} 
 \label{conv-core}
For the convolutions in the denoising block, which vary in size, we have implemented multiple convolution cores operating in parallel. This parallelism ensures the desired latency is achieved while maintaining efficient resource usage and minimizing power consumption. Each convolution core utilizes a 3 \(\times\) 3 systolic array architecture with an input-stationary dataflow. This dataflow is optimized by connecting the convolution core to separate on-chip memory units for weights and activations, ensuring efficient data access and reducing memory access latency.

%% file: results.tex
\subsection{Experimental Setup}

Our experiments involve widely recognized neural network models and datasets. Specifically, we utilize the ImageNet-1k as the pretraining evaluation benchmark, and CIFAR dataset as our transfer learning evaluation benchmarks. We utilize a range of models, including MobileNet-V2 \cite{sandler2018mobilenetv2}, ResNet-18 \cite{DBLP:journals/tip/ZhangZCM017}, and EfficientNet-B0 \cite{tan2019efficientnet}, and DenseNet-121 \cite{huang2017densely}. The pretrained versions of these models were obtained from the Timm library. To simulate the noisy multiply-accumulate operations (MACs), Gaussian noise was introduced to all layers of the models. To mitigate the impact of this noise, we employ the algorithm described in Section \ref{sec:model_arch} to determine the layers for inserting the denoising blocks. In our experiments, the parameter budget \(\eta\) is set to 4\% of the total model parameters. Each denoising block is configured to match the input dimensions of the corresponding input feature tensor, with a bottleneck dimension ratio of 1/4, similar to the configuration in \cite{DBLP:journals/tip/ZhangZCM017}. Once the model is augmented with the denoising modules, we train only the denoising blocks, keeping the backbone of the original model fixed. Additionally, we assume that the \textbf{operations within the denoising blocks are noise-free}. All training experiments were conducted using PyTorch on a NVIDIA A6000 GPU. The denoising blocks were trained for 5 epochs using ADAM optimizer \cite{kingma2014adam} with a learning rate of \(10^{-3}\). For the hardware, we have used 4 parallel convolutional cores and 4 parallel noise cancellation blocks. In all experiment, we set the precision of the weights and activations to 16-bit using fixed-point uniform quantization \cite{jacob2018quantization, azizi2023sensitivity, gholami2022survey}. 

We implemented the denoising block in the digital circuit using High-Level Synthesis (HLS) with Synopsys tools. This implementation is based on 45-nanometer nanGate technology. The operating frequency of the circuit is 500 MHz in our implementation. In our implementation, we utilized 10 parallel convolution cores, as discussed in \ref{conv-core}. For the noise cancellation block, we incorporated 4 parallel denoising lanes to accelerate the denoising process.

\begin{table}
\centering
\caption{Performance of our Denoising Mechanism}
\label{tab:results}
\resizebox{\columnwidth}{!}{
\begin{tabular}{c c c c c c}

\toprule
\multirow{2}{*}{\textbf{Dataset}}   & \multirow{2}{*}{\textbf{Architecture}}    & \textbf{Baseline Model} & \textbf{Noisy Model}         & \textbf{Denoised Model}         &  \multirow{1}{*}{\textbf{Parameter Count}} \\
{}                                  & {}                                        & {\textbf{Accuracy (\%)}}  & \textbf{\textbf{Accuracy (\%)}}             & {\textbf{Accuracy (\%)}}     & {\textbf{Overhead (\%)}} \\
\midrule[\heavyrulewidth]
\multirow{5}{*}{\rotatebox[origin=c]{90}{\textbf{ImageNet}}}    & \multirow{1}{*}{\textbf{ResNet-18}}    &
{71.6}    & {61.5}    & {\textbf{71.5}}    & {0.6} \\

\cmidrule[\heavyrulewidth]{2-6}
{}  & \multirow{1}{*}{\textbf{MobileNetV2}}    &
72.6    & {66.05}    & {\textbf{71.6}}    & {3.2} \\

\cmidrule[\heavyrulewidth]{2-6}
{}  & \multirow{1}{*}{\textbf{EfficientNet-B0}}    & 
77.78    & {0.43}    & {\textbf{73.55}}    & {2.13} \\

\cmidrule[\heavyrulewidth]{2-6}
{}  & \multirow{1}{*}{\textbf{DenseNet-121}}    &
75.45    & {32.10}    & {\textbf{72.06}}    & {3.48} \\

\midrule[\heavyrulewidth]
\multirow{4}{*}{\rotatebox[origin=c]{90}{\textbf{CIFAR-10}}}   & \multirow{1}{*}{\textbf{ResNet-18}}  & 
96.19    & {92.03}    & {\textbf{96.23}}    & {0.67} \\

\cmidrule[\heavyrulewidth]{2-6}
{}  & \multirow{1}{*}{\textbf{MobileNetV2}}    &
93.43    & {88.22}    & {\textbf{93.48}}    & {1.37} \\

\cmidrule[\heavyrulewidth]{2-6}
{}  & \multirow{1}{*}{\textbf{EfficientNet-B0}}    & 
92.56    & {17.38}    & {\textbf{93.1}}    & {2.8} \\

\bottomrule
\end{tabular}}
\end{table}

\subsection{Evaluation Results}
Table \ref{tab:results} presents the performance comparison of our denoised models against their baseline counterparts and the noisy versions (prior to denoising). Introducing Gaussian noise with a mean of 0 and a standard deviation equal to 6\% of the magnitude of the feature map across all layers of ResNet-18, MobileNet-v2 and DesNet-121 results in performance drops of 11.1\%, 6.55\%, and 43.35\%, respectively. By employing our denoising framework and adding less than 4\% of the original model’s total parameters, these performance drops are substantially mitigated, reducing to just 0.1\% for ResNet-18, 1\% for MobileNet-v2, and 3.39\% for DesNet-121. Interestingly, when the same amount of noise is applied to EfficientNet-B0, the performance of the model is severely impacted, with accuracy dropping nearly to zero. This behavior can be attributed to the highly optimized, compressed nature of EfficientNet, which is designed through neural architecture search to achieve state-of-the-art efficiency. As a result, any perturbation to the model can substantially disrupt these finely tuned optimizations. Nevertheless, our denoising blocks demonstrate their effectiveness by recovering the model’s accuracy within just 5 epochs of training.

For the CIFAR transfer learning task, we first fine-tune the pretrained ImageNet-1k models on the CIFAR-10 dataset to establish the baseline. We then follow a procedure analogous to the one used for the ImageNet baseline to obtain the denoised models. As shown in the table, ResNet-18, MobileNetV2, and EfficientNet-B0 experience performance drops of 4.14\%, 5.21\%, and 75.18\% respectively in the presence of noise. Remarkably, our denoising framework fully mitigates this performance degradation, restoring the accuracy of the models while adding less than 3\% additional parameters.

We have integrated our digital denoising block into the MX-CGRA accelerator proposed in \cite{10.1145/3595638}. The cycle count of the convolutional layers in ResNet-18, both before and after the addition of the denoising block, is illustrated in Fig. \ref{fig:latency}. Only the iteration count for the first and twentieth layers has changed, as these are the layers where an extra denoising block has been introduced. A comparison of the total iteration count before and after the denoising demonstrates that the addition of the denoising block increases latency by only 11\% with the average power overhead of 1.78 mW measured by the Synopsys Design Compiler (DC). 
\begin{table}
\centering
\caption{Effect of the Noise \(\sigma\) on the our denoising framework performance}
\begin{flushleft} \scriptsize
 \(\sigma\) is expressed as a percentage of the magnitude of the corresponding signal. 
\end{flushleft}
\label{tab:sigma}
\resizebox{\columnwidth}{!}{
\begin{tabular}{c c c c c c}

\toprule
\multirow{2}{*}{\textbf{Dataset}}   & \multirow{2}{*}{\textbf{Architecture}}    & \textbf{Baseline Model}  &  \multirow{2}{*}{\textbf{Gaussian Noise \(\sigma\) (\%)}} & \textbf{Noisy Model}         & \textbf{Denoised Model}         \\
{}                                  & {}                                        & {\textbf{Accuracy (\%)}}  &  {} & \textbf{\textbf{Accuracy (\%)}}             & {\textbf{Accuracy (\%)}}      \\
\midrule[\heavyrulewidth]
\multirow{5}{*}{\rotatebox[origin=c]{90}{\textbf{ImageNet}}}    & \multirow{1}{*}{\textbf{ResNet-18}}    &
{71.6}    & {2.0}    & {69.2}    & {71.58} \\

\cmidrule[\heavyrulewidth]{2-6}
{}  & \multirow{1}{*}{\textbf{ResNet-18}}    &
71.6    & {4.0}    & {65.4}    & {71.55} \\

\cmidrule[\heavyrulewidth]{2-6}
{}  & \multirow{1}{*}{\textbf{ResNet-18}}    & 
71.6    & {6.0}    & {51.5}    & {71.5} \\

\cmidrule[\heavyrulewidth]{2-6}
{}  & \multirow{1}{*}{\textbf{ResNet-18}}    &
71.6    & {8.0}    & {53.55}    & {71.1} \\

\bottomrule
\end{tabular}}
\end{table}

\subsection{Ablation Study}
In this section, we evaluate the effectiveness of our framework across different noise levels. We introduce Gaussian noise to ResNet-18 with varying noise strengths, characterized by different values of \(\sigma\), and apply our denoising process. The results are presented in Table \ref{tab:sigma}. When \(\sigma\) is set to 2\% of the feature map magnitude, the model experiences a minor accuracy drop of 2.4\%. However, as \(\sigma\) increases to 8\% of the feature map magnitude, the accuracy drop becomes substantial, reaching 18.05\%. Notably, our method effectively recovers the accuracy, reducing the gap to just 0.5\%. These results demonstrate the robustness and effectiveness of our framework in mitigating performance degradation across a wide range of noisy conditions.

%% file: conclusion.tex
In conclusion, our method presents an innovative denoising framework that effectively preserves the accuracy of DNNs in mixed-signal accelerators despite the noise introduced by process and temporal variations. Notably, this approach eliminates the need to retrain the entire model while maintaining low latency and minimal power consumption overhead. Additionally, our framework is versatile and can be applied to any mixed-signal DNN accelerator, enhancing the robustness of these models.